\documentclass[runningheads]{llncs}
\usepackage[T1]{fontenc}
\usepackage{graphicx,verbatim}
\usepackage{amsmath}
\usepackage{amsfonts}
\usepackage{booktabs}
\usepackage{multirow}
\usepackage{makecell}
\usepackage{placeins}
\usepackage{hyperref}  
\usepackage{xcolor}
\usepackage{booktabs, makecell} % in preamble

\begin{document}
\title{ZODIAC: Zero-shot Octree-based Diffusion for Anatomical Completion}

\author{Miruna-Alexandra Gafencu\inst{1, 2,3}\thanks{These authors contributed equally to this work.}\and
Vlad Bratulescu\inst{1}\textsuperscript{*}
 \and
 Yordanka Velikova\inst{1, 2}
 \and
 Mohammad Farid Azampour\inst{1, 2} 
 \and
 Nassir Navab\inst{1, 2}
 }

\authorrunning{M.-A. Gafencu et al.}
\institute{Chair for Computer Aided Medical Procedures (CAMP) \\ Technical University of Munich, Boltzmannstr. 3, 85748 Garching, Germany 
\email{\{miruna.gafencu, vlad.bratulescu, dani.velikova, mf.azampour, nassir.navab\}@tum.de}
 \and
Munich Center for Machine Learning (MCML), Munich, Germany
\and Konrad Zuse School of Excellence in Reliable AI (relAI), Germany}

\maketitle              % typeset the header of the contribution
\sloppy
\begin{abstract}
Recovering the full 3D spine anatomy from intraoperative ultrasound is an ill-posed inverse problem, as the complete structure must be inferred from incomplete and noisy observations. Acoustic occlusions and limited field of view create large unobserved regions, while view-dependent artifacts lead to variability in expert annotations of the visible anatomy. Current supervised ultrasound shape completion methods rely on synthetically generated incomplete–complete paired data to learn conditional mappings under a predefined distribution of simulated occlusions. However, real intraoperative occlusions do not necessarily follow this distribution, which can limit generalization to patient data. As a result, accurate and robust completion from noisy partial observations remains an unsolved problem. We propose a zero-shot shape completion framework that reconstructs the entire lumbar spine from partial ultrasound observations without relying on simulated training data. To accommodate unseen and irregular patterns of missing structures, we introduce blended completion, a mechanism that integrates the learned anatomical prior with incoming partial geometry at inference time. The method learns a generative diffusion prior over full anatomical shapes represented in an adaptive octree structure, enabling efficient modeling of the complete spine in a single forward pass. 
Validation on phantom and volunteer data shows that decoupling completion from a predefined corruption distribution improves generalisation under real occlusions, outperforming a fully supervised variant by $22\%$ on HD95 completion error. Code and data are available at https://github.com/miruna20/ZODIAC.


% Authors must provide keywords and are not allowed to remove this Keyword section.

\end{abstract}

\section{Introduction}  
Ultrasound (US) is increasingly used to guide spinal interventions such as epidural injections and pedicle screw placements~\cite{li2023robot,li2024ultrasound}, owing to its real-time feedback, absence of ionising radiation, and low cost. Yet performing these procedures is cognitively challenging, as it requires the clinician to maintain an accurate mental map of the underlying 3D spinal anatomy; information that ultrasound alone cannot directly provide. Since bone strongly reflects ultrasound waves and casts deep acoustic shadows, the spine is only partially visible. Therefore, the clinician must mentally reconstruct the complete 3D structure based on internalised anatomical knowledge.

From a computational perspective, this reconstruction corresponds to an ill-posed inverse problem: the full 3D anatomy must be inferred from incomplete and noisy observations. Large occluded regions result in missing structural information, while view-dependent artifacts degrade the reliability of the visible anatomy. In practice, partial geometry is typically extracted through ultrasound segmentation, yet segmentation quality varies across operators and acquisition conditions. As a result, shape completion methods must operate robustly on inputs that are both incomplete and uncertain.

Recent ultrasound shape completion methods \cite{gafencu2024shape,gafencu2025us} attempt to address this problem by learning conditional mappings between partial observations and complete 3D anatomy. Due to the limited availability of paired ultrasound–CT datasets, these approaches typically rely on synthetically generated incomplete–complete pairs derived from CT data. While this enables supervised learning, it imposes a predefined distribution of simulated occlusions and segmentation artifacts. However, real intraoperative missing structures do not necessarily follow this predefined corruption process. Occlusions vary with probe pose, anatomy, and patient-specific factors, and segmentation errors are irregular and unpredictable. As a result, models trained under synthetic corruption assumptions may struggle to generalize when confronted with noisy or structurally incomplete patient data. Beyond supervision, structural design choices further constrain existing approaches. They represent anatomy as point clouds, where increasing geometric detail requires a proportional increase in point density and memory consumption. In addition, point-based representations lack an explicit surface formulation and require post-processing to generate meshes. Most importantly, prior work completes vertebrae independently, ignoring inter-vertebral geometric context. The lumbar spine, however, is an articulated structure whose global curvature and relative vertebral pose are inherently coupled. Independent completion risks producing geometrically inconsistent reconstructions. Given  these current limitations, our goal is to provide a robust whole-spine shape completion framework, that removes the dependence on paired training data, represents the complex spine anatomy efficiently and  produces a complete reconstruction directly usable for surgical visualization. 

We therefore seek a suitable representation method as well as a generative framework capable of learning a prior over the complex spine geometry. 
Octrees are a hierarchical, sparse representation that concentrates computation and memory on occupied regions. OctNet~\cite{riegler2017octnet} and O-CNN~\cite{wang2017cnn} showed they support deep volumetric learning, for tasks such as 3D classification and segmentation, at resolutions dense grids cannot reach. 
Diffusion models learn a data distribution by reversing a gradual noising process~\cite{ho2020denoising,rombach2022high} and have become a leading approach to high-fidelity generation, first for images~\cite{rombach2022high} and increasingly for 3D shapes, where they form strong priors over plausible geometry~\cite{cheng2023sdfusion}. In our case this is suitable, as the prior is learned from complete examples alone. OctFusion~\cite{xiong2025octfusion} couples the two, generating high-resolution shapes on an octree representation at low computational and memory cost. Its backbone has since been used for fine-grained generation~\cite{gao2025hieroctfusion}, semantic scene generation~\cite{zhang2025octree}, and autoregressive generation~\cite{wei2025octgpt}.

Building on this octree-diffusion backbone we propose ZODIAC: Zero-shot Octree-based Diffusion for Anatomical Completion, the first approach for whole-spine 3D shape completion from partial ultrasound observations. Rather than relying on point cloud representations, ZODIAC operates on the adaptive octree representation. This hierarchical structure concentrates model capacity near bone surfaces to capture anatomical details while efficiently encoding the spine's global structure. A central aspect of ZODIAC is its zero-shot formulation: it requires no paired incomplete–complete training data and performs completion training-free at inference on an unconditionally trained prior. Rather than learning under a predefined synthetic corruption distribution, we introduce a blended completion mechanism that integrates this prior with incoming partial geometry at inference time, enabling reconstruction under irregular and previously unseen patterns of missing structures. 
    
We validate the proposed framework on phantom and volunteer datasets, and perform ablation studies and baseline comparisons, demonstrating especially stable performance under noisy and structurally incomplete observations.

\section{Methodology}
\label{sec:method}

\subsection{Problem Formulation}

Let $\mathcal{S} \subset \mathbb{R}^3$ denote the complete anatomy and let $\mathcal{P} \subset \mathcal{S}$ be a partial observation corrupted by missing regions and measurement noise, such that $\mathcal{P}$ covers only a subset of $\mathcal{S}$ with no guarantee on the spatial distribution of the missing structure. Shape completion amounts to inferring the full shape $\mathcal{S}$ from $\mathcal{P}$, i.e.\ approximating the posterior $p(\mathcal{S} \mid \mathcal{P})$.
Rather than learning this posterior directly from paired data, we decompose it into a learned prior $p(\mathcal{S})$ over complete anatomies and an inference-time mechanism that conditions generation on $\mathcal{P}$. This enables a zero-shot constraint: at training time, only a set of $N$ complete shapes $\{\mathcal{S}_i\}_{i=1}^{N}$ is needed, with no required access to paired $(\mathcal{P}_i, \mathcal{S}_i)$ data. The partial observation $\mathcal{P}_i$ is incorporated only at inference time. 
% ---------------------------------------------------------------
\subsection{Octree Representation and Generative Prior}
\label{subsec:repandprior}
  \begin{figure}
 \includegraphics[width=\textwidth]{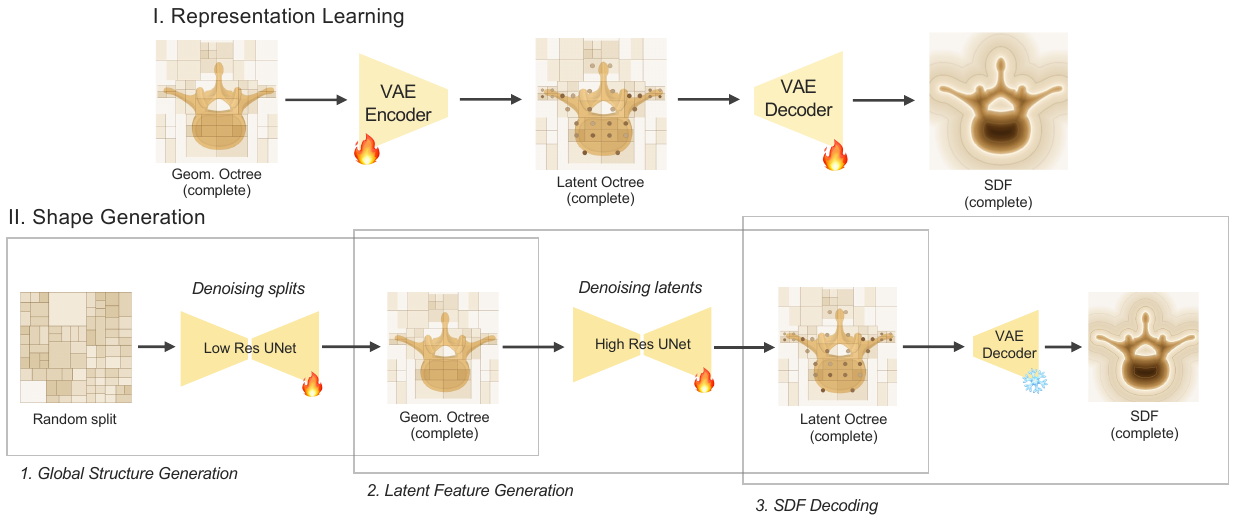}
 \caption{\textbf{Training pipeline.} In Phase I (Representation Learning), a VAE encoder maps a geometric octree to a latent octree, which is reconstructed into a continuous SDF. In Phase II (Shape Generation), a two-stage denoising diffusion probabilistic model (DDPM) generates shapes from noise: a low-resolution UNet first denoises octree splitting signals to produce a geometric octree, followed by a high-resolution UNet that denoises the corresponding latent features. Finally, the resulting latent octree is decoded into a final SDF.} \label{fig:training}
\end{figure}

In our specific problem, we aim to represent and complete the lumbar spine, a complex, multi-part anatomy. We therefore employ octrees as a compact, adaptive 3D representation that concentrates resolution near anatomical surfaces while avoiding the memory cost of dense voxel grids.\\

\subsubsection{Shape Representation.} Based on~\cite{xiong2025octfusion} we adapt the octree diffusion framework. First we encode a 3D shape $\mathcal{S}_i$ as a \emph{geometric octree} 
$\mathcal{O}$ of depth $D$, constructed by recursively subdividing 
the cubic bounding volume $\Omega$ of its point cloud representation $\mathbf{S}_i \in \mathbb{R}^{M \times 3}$
and retaining only nodes containing surface points. Nodes without children form the set 
of leaf nodes $\mathcal{L}_i$, each spanning a cube of side length 
$\ell(\Omega)/2^d$, where $d \leq D$ is the node depth.

\subsubsection{Representation Learning.} To enable diffusion modeling in 
a compact latent space, we first compress the geometric octree $\mathcal{O}$ into a lower-dimensional latent octree $\mathcal{Z}$ using a variational autoencoder 
(VAE)~\cite{wang2022dual} as displayed in 
Fig.~\ref{fig:training} Phase I. Each leaf node stores a $C$-dimensional feature vector $\mathbf{z}_v \in \mathbb{R}^C$, encoding surface geometry. The decoder simultaneously recovers the octree structure by predicting binary split decisions at each depth level, and decodes the latent features into a continuous signed-distance field (SDF). The final surface is extracted by applying marching cubes. We train the VAE with a combination of cross-entropy loss on split 
predictions, SDF reconstruction loss, and KL divergence regularisation, and freeze 
it after training.

\subsubsection{Generative Prior.} We train a two-stage denoising diffusion 
probabilistic model (DDPM)~\cite{ho2020denoising}, shown in 
Fig.~\ref{fig:training} Phase II, to learn the distribution over complete shapes 
$p(\mathcal{S})$ from the training set. Generation proceeds coarse-to-fine. \\

\textbf{The LR Diffusion Stage}: A low-resolution UNet first denoises the base-depth split status of the octree, capturing global spine structure and inter-vertebral configuration. Up to the base depth $D_{\text{base}}$ the octree is full, i.e.\ every node exists, so its structure reduces to a uniform split grid of resolution $2^{D_{\text{base}}} \times 
2^{D_{\text{base}}} \times 2^{D_{\text{base}}}$. For each grid cell, the LR representation stores the binary split status of its eight children. The status can be active, indicating there is a split or inactive indicating no split. A node is subdivided whenever at least one of its children split states is active. The active child octants are then refined once more according to their predicted split states, recursively growing the coarse octree from depth $D_{\text{base}}$ to $D_{\text{base}} + 2$. For the diffusion process these binary signals are stacked into a dense grid with the octant split decisions per node 
${\mathbf{G}}_0 \in \{-1,1\}^{8 \times 2^{D_{\text{base}}} \times 2^{D_{\text{base}}} \times 2^{D_{\text{base}}}}$, 
where active and inactive splits are encoded as $+1$ and $-1$.\\

\textbf{The HR Diffusion Stage}: Conditioned on the resulting structure, a 
high-resolution graph UNet then denoises the per-node latent codes 
$\mathbf{z}_v$ and recovers fine surface detail within the coarse envelope. The final 
surface is obtained by decoding $\mathcal{Z}$ through the frozen VAE decoder 
and running marching cubes on the resulting SDF.

\subsection{Zero-shot Shape Completion}
\label{sec:completion}
Given the trained generative model, we formulate shape completion as a constrained generation problem: sample from the prior $p(\mathcal{S})$ 
while respecting the partial observation $\mathcal{P}$. Since paired $(\mathcal{P}_i, \mathcal{S}_i)$ data is unavailable, we 
propose a training-free, zero-shot completion strategy that operates 
directly on the unconditionally trained model by blending the partial  observation into each reverse diffusion step at test time.

\subsubsection{Partial Observation Rasterization.}
\label{subsubsec:partialobservrasterization}
Given the partial observation $\mathcal{P}$, provided as a point cloud $\mathbf{P} \in \mathbb{R}^{K \times 3}$, we 
build an octree from its points and extract the base-depth split signal with 
the same procedure used for complete shapes, yielding the rasterized  
grid $\tilde{\mathbf{G}} \in \{-1,1\}^{8 \times 2^{D_{\text{base}}} \times 
2^{D_{\text{base}}} \times 2^{D_{\text{base}}}}$ representing the partial observation. We then define a binary mask $\mathbf{m} \in \{0,1\}^{8 \times 
2^{D_{\text{base}}} \times 2^{D_{\text{base}}} \times 2^{D_{\text{base}}}}$ 
aligned with $\tilde{\mathbf{G}}$, where $\mathbf{m} = 1$ 
at the split entries the partial observation marks as occupied and $\mathbf{m} = 0$ otherwise. The mask is 
one-sided by design: it anchors only positively observed geometry, while 
unobserved entries ($\mathbf{m} = 0$) are left free for the prior to 
complete. This reflects the nature of ultrasound occlusion, where the 
absence of an observed surface does not confirm empty space but  
indicates an unscanned or shadowed region.

\begin{figure}
 \includegraphics[width=\textwidth]{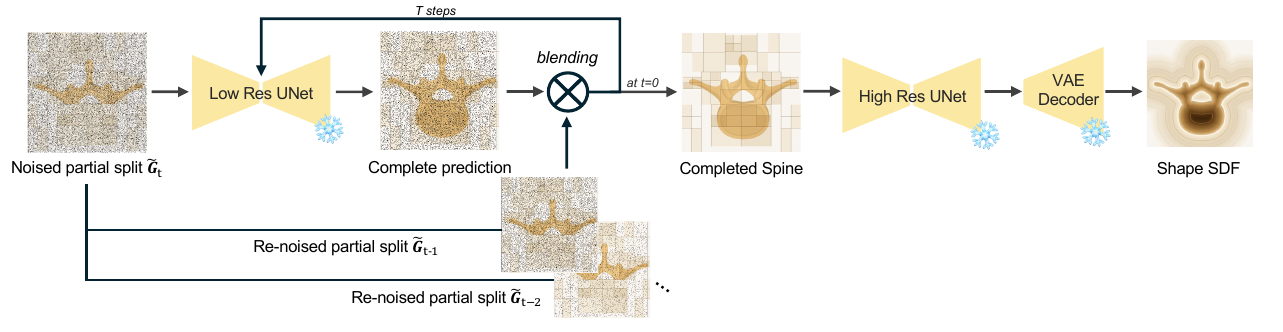}
 \caption{\textbf{Zero shot shape completion mechanism.} Within the LR diffusion stage, the partial input is re-noised and blended back at each denoising step, constraining the generation to remain consistent with the observed structure. The resulting geometric octree is then passed to the HR diffusion stage to denoise the latent features. Finally, the frozen VAE decoder produces the complete shape SDF. } \label{fig:inference}
\end{figure} 

\subsubsection{Blended Completion.} Inspired by~\cite{avrahami2022blendeddiffusion} and~\cite{lugmayr2022repaint}, we introduce the blended completion mechanism illustrated in Fig.~\ref{fig:inference}. Rather than starting the LR reverse diffusion process from pure noise, we initialize it from a noised version of the partial observation:
\begin{equation}
    \mathbf{G}_T = \sqrt{\bar{\alpha}_T}\,\tilde{\mathbf{G}} + 
    \sqrt{1 - \bar{\alpha}_T}\,\boldsymbol{\epsilon}, 
    \quad \boldsymbol{\epsilon} \sim \mathcal{N}(\mathbf{0}, \mathbf{I}),
    \label{eq_1}
\end{equation}
where $\mathbf{G}_T$ is the noised grid at the final diffusion step $T$, $\tilde{\mathbf{G}}$ is the rasterized partial grid without any noise added, and $\boldsymbol{\epsilon}$ is standard Gaussian 
noise. The coefficient $\bar\alpha_t$ is the cumulative signal coefficient of the variance schedule, controlling the signal-to-noise ratio. At each subsequent reverse step, we first obtain the unconstrained DDPM estimate $\hat{\mathbf{G}}_t$ by denoising $\mathbf{G}_{t+1}$. Then, the observed regions are replaced with the corresponding re-noised version
$\tilde{\mathbf{G}}_t =\sqrt{\bar{\alpha}_t}\,\tilde{\mathbf{G}} 
+ \sqrt{1-\bar{\alpha}_t}\,\boldsymbol{\epsilon}$ as follows:

\begin{equation}
    \mathbf{G}_t = \mathbf{m} \odot \tilde{\mathbf{G}}_t 
    + (\mathbf{1} - \mathbf{m}) \odot \hat{\mathbf{G}}_t, 
    \qquad t = T-1, \ldots, 0, 
    \qquad \mathbf{G}_T \text{ as in Eq.~\ref{eq_1}}.
    \label{eq_2}
\end{equation}

where $\odot$ is the elementwise product, $\boldsymbol{\epsilon}$ is resampled each step and $\tilde{\mathbf{G}}_t$ denotes the partial re-noised at level t. Iterating this update until $t = 0$ yields the completed low-resolution grid 
$\mathbf{G}_0$, which is passed to the HR stage for surface refinement.
Re-noising ensures the observed regions remain statistically consistent 
with the training distribution, avoiding the distribution shift that 
would arise from injecting the clean $\tilde{\mathbf{G}}$ directly. We ablate the re-noising mechanism within the ZODIAC-clean method.  

As shown in Fig.~\ref{fig:inference}, we apply blending exclusively 
at the LR stage. The LR stage determines global spine topology, such as
vertebral arrangement and overall curvature, which is precisely what 
$\mathbf{P}$ constrains. At each denoising step, observed entries are 
anchored to $\tilde{\mathbf{G}}$ while the remaining unobserved entries evolve freely 
under the learned prior, completing the missing structure. Due to
segmentation noise, $\mathbf{P}$ cannot reliably constrain fine surface 
details; the HR stage instead refines geometric details uniformly across observed and unobserved regions alike, conditioned on the globally consistent structure produced by the LR stage.

\section{Experiments}

\subsection{Datasets}
\subsubsection{Training Data.}
For 91 lumbar spine meshes from VerSe20~\cite{sekuboyina2021verse} we apply deformation augmentation as described in~\cite{azampour2024deformations} to account for different spine curvatures. Together with 322 lumbar spine meshes extracted from the TotalSegmentator dataset \cite{wasserthal2023totalSegmentator} we create our training set of complete shapes.

\subsubsection{Evaluation Data.}
For evaluation, we include the acquisitions of two anthropomorphic lumbar spine phantoms~\cite{gafencu2025us} and six scans from three volunteers in a study conducted at Balgrist Hospital~\cite{cavalcanti2025large}. We refer to the latter as the "Balgrist dataset", the only publicly available dataset of paired US-CT acquisitions of volunteers. Each subject contributes one handheld and one robotic ultrasound acquisition. To obtain a point cloud, the lumbar spine surface, manually annotated by a physician and included in the published dataset, is first extracted. It is then manually registered to the corresponding CT annotation by one operator and verified by a second, both experienced in spinal US. Only scans with clear anatomical anchors and identifiable vertebral levels are selected (n=6), and registration is performed such that these landmarks remain consistently visible across both modalities. While registration of the occluded structures in the ultrasound scan cannot be directly validated, any residual misalignment is shared across all compared methods and therefore does not affect relative comparisons. Notably, two of the handheld scans were acquired in a manner closely resembling clinical use, making them particularly valuable for robustness estimation.

\subsubsection{Training Setup.}
For shape representation we rasterize each lumbar spine into a $256^3$ signed distance field and train the model in
three stages. The octree VAE is trained first for 900 epochs using AdamW with a learning rate of $10^{-3}$, optimising an SDF regression loss with a KL regulariser (weight 0.1). The two diffusion stages are then trained successively with AdamW at a learning rate of $2\times10^{-4}$. The coarse (LR) diffusion model is trained for 1500 epochs, and the refinement (HR) model, initialised from the LR checkpoint, for 300 epochs. All stages use a batch size of 1, an exponential moving average of the network weights with decay 0.999, and a fixed random seed for reproducibility.
Training was performed on an NVIDIA GeForce RTX 4080 and uses a total of 504 spines. At inference, shapes are sampled with 200 diffusion steps. The shape completion using the trained model requires approximately 8 seconds per spine.

\subsection{Evaluation}

\subsubsection{Metrics.} We report Chamfer Distance (CD), F1-score at 1\% threshold and 95\% Hausdorff Distance (HD95). Per-vertebra metrics are obtained by segmenting the predicted spine mesh into L1–L5 based on proximity to ground truth levels. 

\subsubsection{Baseline}
We compare to Shape Completion in the Dark~\cite{gafencu2024shape}, a 
baseline which independently completes each lumbar vertebra from
partial ultrasound point clouds. To enable whole-spine metric evaluation, we merge the five individually completed vertebrae into a single point cloud.

\subsubsection{Trained with Pairs Variant (TP-ODIAC)} 
\label{subsec:partial_cond}
We implement TP-ODIAC, the supervised variant of ZODIAC. This method requires paired partial-complete training data and is expected to serve as an upper bound on completion accuracy. Given a partial point cloud $\mathbf{P} \in \mathbb{R}^{K \times 3}$ and its corresponding complete shape point cloud $\mathbf{S} \in \mathbb{R}^{M \times 3}$,  $K < M$,  we construct octrees and extract their base-depth split signal $\tilde{\mathbf{G}}$ and ${\mathbf{G}_0}$ as described in Sec.~\ref{subsubsec:partialobservrasterization}, where  ${\mathbf{G}_0}$ serves as the diffusion target and $\tilde{\mathbf{G}}$ as the conditioning signal. Compared to ZODIAC, $\tilde{\mathbf{G}}$ is now channel-concatenated with the current ${\mathbf{G}}_t$ and passed to the LR UNet. $\tilde{\mathbf{G}}$ is passed clean and unmodified at every timestep $t$, independently of the noise level. This provides the UNet with a constant observation signal throughout the entire denoising trajectory. The same conditioning is applied identically at training and inference time. To train TP-ODIAC, we simulate ultrasound-consistent partial point clouds with realistic artifacts and diverse spine curvatures following ~\cite{gafencu2024shape}.

\subsubsection{Zero-shot Variant with Hard Observation Constraints (ZODIAC-clean)}
As an ablation to the noising mechanism of blended completion, we additionally implement ZODIAC-clean, in which the partial observation is treated as a hard boundary condition. The sampling trajectory is initialized with the clean observation at the observed voxels and pure Gaussian noise elsewhere, and at every denoising step the observed voxels are pinned directly to the noise-free partial, $\mathbf{G}_t = \mathbf{m} \odot \tilde{\mathbf{G}} + (\mathbf{1} - \mathbf{m}) \odot \hat{\mathbf{G}}_t$, where $\mathbf{m}$ is the observation mask. This enforces exact agreement with the input throughout sampling and restricts the diffusion prior to completing only the unobserved regions.

\section{Results and Discussion}

\begin{figure}
 \includegraphics[width=\textwidth]{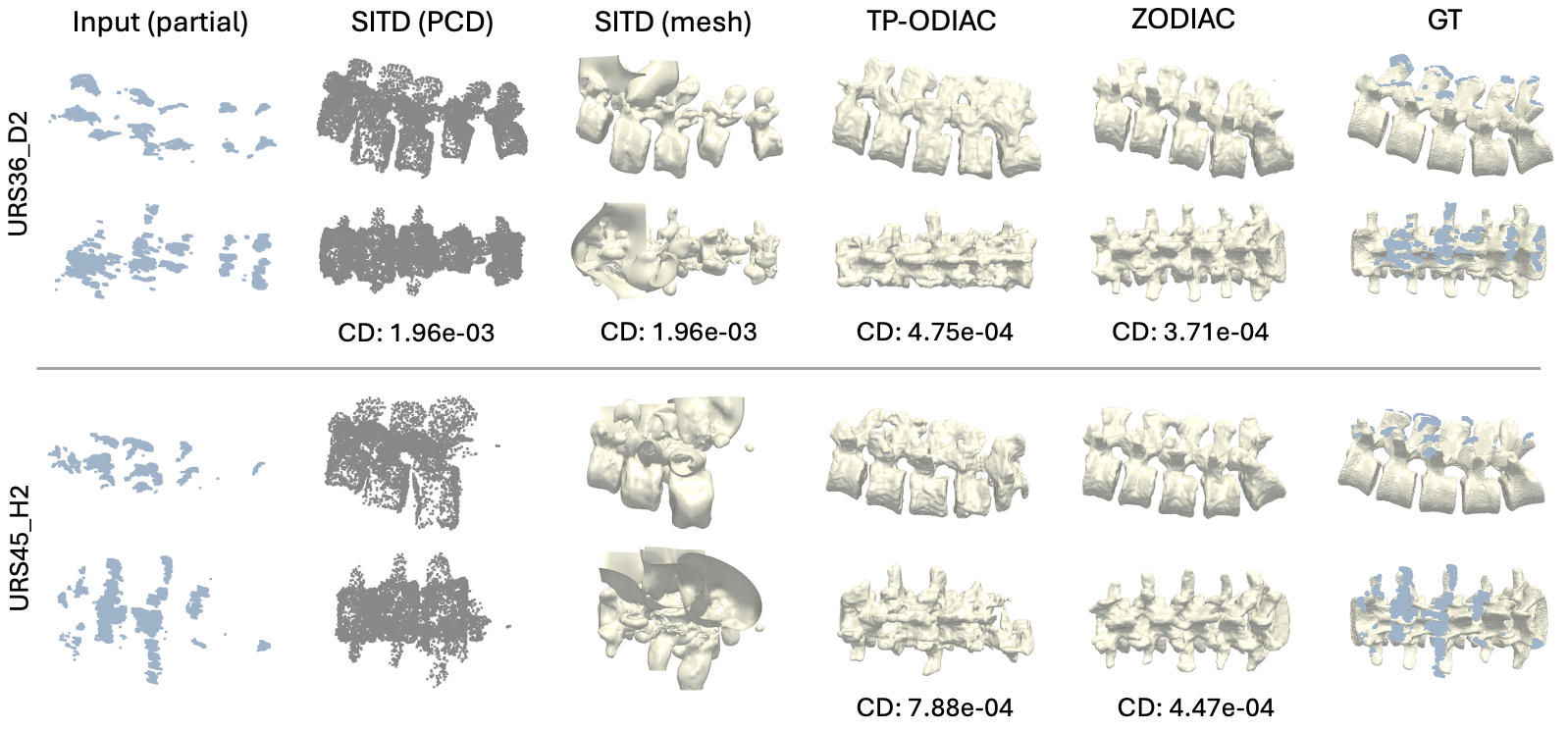}
 \caption{\textbf{Qualitative analysis of completed meshes.} ZODIAC produces anatomically and geometrically coherent completions compared to SITD (baseline) and TP-ODIAC (fully supervised), particularly on volunteer acquisitions, where baselines yield fragmented or collapsed surfaces.} \label{fig:qualitativeResults}
\end{figure} 

\subsection{Comparison with State of the Art}
We first compare ZODIAC against SITD,  state-of-the-art (SOTA) which requires training pairs. We display the results in Table~\ref{tab:phantom_balgrist_combined}. SITD performs shape completion independently per vertebra and depends on point clouds with sufficient coverage and low noise for each vertebra. We therefore report SITD whole-spine metrics only where all five vertebrae are reconstructed. 
On phantom acquisitions, ZODIAC performs on par with SITD at the whole-spine level, while  vertebra-wise it improves CD by $21\%$ and HD95 by $14\%$. This gap widens on volunteer data, where ZODIAC improves vertebra-wise CD by $70\%$ and HD95 by $47\%$ as well as the scan-to-scan variability from $\pm28.42$ to $\pm3.60$, reflecting the sensitivity of SITD to the noise and incomplete coverage inherent in real ultrasound acquisitions. These results demonstrate that zero-shot whole-spine joint modeling can outperform the SOTA supervised baseline without requiring any paired 
training data, while remaining more robust under the challenging conditions of in-vivo ultrasound. 

\subsection{Comparison with supervised variant (TP-ODIAC)}
We compare ZODIAC against TP-ODIAC across 
both phantom and volunteer data. The results are displayed in Table \ref{tab:phantom_balgrist_combined}. This comparison is intentionally 
asymmetric: TP-ODIAC has access to task-specific paired training data, 
while ZODIAC relies solely on complete shape priors learned from 
publicly available datasets, requiring no 
paired incomplete--complete examples. As expected for a supervised upper bound, TP-ODIAC leads on the phantom, where acquisitions are clean and in-distribution. On volunteer data the gap closes to near-parity: ZODIAC is within 3\% of TP-ODIAC on whole-spine CD and matches it on HD95, without any paired supervision, though F1-score still trails by 15\%. Moreover, ZODIAC produces more consistent completions across subjects with different coverage regimes, with a between-subject standard deviation in CD roughly 80\% lower than TP-ODIAC on the whole-spine. 

We additionally compare on difficult cases, which are  volunteer acquisitions where the spine surface is only partially visible and one or more vertebrae are entirely absent from the partial input. On these cases, ZODIAC 
outperforms TP-ODIAC whole-spine in 2 out of 3 metrics with CD by 19\% and HD95 by 22\%.  F1-score remains 11\% behind, indicating that ZODIAC's outlier suppression drives the CD and HD95 gains but not F1, which counts true and false positives within a fixed distance. Qualitatively,  Fig. \ref{fig:qualitativeResults} shows that TP-ODIAC fails to reconstruct important landmarks such as the lateral processes when they are not present in the partial input. In contrast, ZODIAC consequently generates an anatomically coherent spine with the expected landmarks. This suggests a key 
limitation of paired supervised approaches. When test-time observations deviate
from the training distribution, the prior-driven completion of ZODIAC degrades less than a model fit to that distribution.

\subsection{Ablation studies - Comparison with ZODIAC-clean}
To isolate the contribution of the per-step blending, we compare ZODIAC against ZODIAC-clean, which anchors to the clean partial observation at every denoising step.
As displayed in Table~\ref{tab:phantom_balgrist_combined} ZODIAC-clean degrades all three metrics across every setting: whole-spine CD rises by $15-26\%$, HD95 by $13-16\%$, and F1 drops by $5-11\%$, with the largest losses on the difficult subset (CD $+26\%$, HD95 $+16\%$). The per-vertebra evaluation shows the same pattern (CD $+18-27\%$). These results demonstrate that anchoring to the clean observation injects a conditioning signal mismatched to the current noise level, and that re-noising at each denoising step benefits more accurate completion. 

\begin{table*}[t]
  \centering
  \caption{Quantitative evaluation on Phantom (n=2) and Balgrist volunteer data.
  Balgrist (all) includes 6 subjects for our methods and 3 for SITD; Balgrist (difficult) is a
  subset of 3 subjects (URS36\_D2, URS45\_H2, URS45\_R2), and SITD reconstructs only one of these.
  CD scaled by $10^4$ (normalised units); F1 is the F-score at the evaluation threshold (higher is better); HD95 in mm.
  All values for our methods are reported as mean $\pm$ standard deviation, where each subject's metric is first averaged over 10   
  stochastic samples and the mean and standard deviation are then taken across  
  subjects. 
  \textbf{Eval.:} WS = Whole Spine, VW = Vertebra-wise.}
  \label{tab:phantom_balgrist_combined}
  \footnotesize
  \setlength{\tabcolsep}{3pt}
  \renewcommand{\arraystretch}{1.05}
  \newcommand{\mstd}[2]{\shortstack[c]{$#1$\\{\scriptsize$\pm #2$}}}
  \newcommand{\mnostd}[1]{\shortstack[c]{$#1$\\{\scriptsize --}}}
  \newcommand{\wsvwcmidrule}{\cmidrule(lr){3-5}\cmidrule(lr){6-8}\cmidrule(lr){9-11}}
  \resizebox{0.95\textwidth}{!}{%
  \begin{tabular}{llccccccccc}
  \toprule
  \multirow{2}{*}{\textbf{Method}} & \multirow{2}{*}{\textbf{Eval.}} &
  \multicolumn{3}{c}{\textbf{Phantom}} &
  \multicolumn{3}{c}{\textbf{Balgrist (all)}} &
  \multicolumn{3}{c}{\textbf{Balgrist (difficult)}} \\
  \cmidrule(lr){3-5}\cmidrule(lr){6-8}\cmidrule(lr){9-11}
  & &
  \textbf{CD$\downarrow$} & \textbf{F1$\uparrow$} & \textbf{HD95$\downarrow$} &
  \textbf{CD$\downarrow$} & \textbf{F1$\uparrow$} & \textbf{HD95$\downarrow$} &
  \textbf{CD$\downarrow$} & \textbf{F1$\uparrow$} & \textbf{HD95$\downarrow$} \\
  \midrule
  \multirow{2}{*}{\makecell[l]{SITD \\ (baseline, supervised)}}
  & WS
  & \mstd{5.38}{0.86}   & \mstd{0.362}{0.029} & \mstd{7.53}{0.75}
  & \mstd{14.94}{3.35}  & \mstd{0.257}{0.018} & \mstd{16.11}{1.42}
  & \mnostd{19.60}      & \mnostd{0.242}      & \mnostd{18.00} \\
  \wsvwcmidrule
  & VW
  & \mstd{15.77}{4.00}  & \mstd{0.193}{0.029} & \mstd{8.25}{1.21}
  & \mstd{52.17}{28.42} & \mstd{0.119}{0.027} & \mstd{13.96}{4.54}
  & \mstd{69.02}{29.87} & \mstd{0.119}{0.017} & \mstd{15.90}{4.30} \\
  \midrule
  \multirow{2}{*}{\makecell[l]{TP-ODIAC \\ (ours, supervised)}}
  & WS
  & \mstd{4.23}{1.61}   & \mstd{0.411}{0.103} & \mstd{6.11}{1.24}
  & \mstd{5.49}{2.46}   & \mstd{0.434}{0.060} & \mstd{7.78}{3.37}
  & \mstd{7.35}{2.15}   & \mstd{0.398}{0.051} & \mstd{10.26}{3.19} \\
  \wsvwcmidrule
  & VW
  & \mstd{9.69}{4.93}   & \mstd{0.240}{0.073} & \mstd{5.70}{1.66}
  & \mstd{14.78}{13.92} & \mstd{0.231}{0.047} & \mstd{7.02}{3.92}
  & \mstd{19.93}{17.49} & \mstd{0.211}{0.040} & \mstd{8.51}{4.57} \\
  \midrule
  \multirow{2}{*}{\makecell[l]{ZODIAC \\ (ours, unsupervised)}}
  & WS
  & \mstd{5.22}{0.98}   & \mstd{0.368}{0.038} & \mstd{7.37}{0.70}
  & \mstd{5.64}{0.47}   & \mstd{0.371}{0.023} & \mstd{7.74}{0.45}
  & \mstd{5.99}{0.44}   & \mstd{0.353}{0.020} & \mstd{8.02}{0.48} \\
  \wsvwcmidrule
  & VW
  & \mstd{12.44}{3.80}  & \mstd{0.213}{0.031} & \mstd{7.08}{1.32}
  & \mstd{15.53}{3.60}  & \mstd{0.198}{0.022} & \mstd{7.37}{0.99}
  & \mstd{16.38}{3.73}  & \mstd{0.188}{0.016} & \mstd{7.67}{1.14} \\
  \midrule
  \multirow{2}{*}{\makecell[l]{ZODIAC-clean \\ (ablation, unsupervised)}}
  & WS
  & \mstd{6.02}{1.70}   & \mstd{0.348}{0.046} & \mstd{8.36}{1.23}
  & \mstd{6.97}{0.74}   & \mstd{0.332}{0.012} & \mstd{8.93}{0.56}
  & \mstd{7.54}{0.64}   & \mstd{0.328}{0.012} & \mstd{9.31}{0.53} \\
  \wsvwcmidrule
  & VW
  & \mstd{14.63}{5.58}  & \mstd{0.200}{0.039} & \mstd{7.78}{1.71}
  & \mstd{19.28}{3.25}  & \mstd{0.175}{0.015} & \mstd{8.60}{0.96}
  & \mstd{20.72}{2.94}  & \mstd{0.176}{0.014} & \mstd{9.05}{0.85} \\
  \bottomrule
  \end{tabular}%
  }
  \end{table*}
\FloatBarrier

\section{Conclusion}
In this work, we introduced ZODIAC, a zero-shot framework for whole-spine 3D shape completion from partial ultrasound observations. By learning a generative prior over complete anatomies and applying blended completion at test time, ZODIAC requires no paired incomplete–complete training data. Crucially, while supervised approaches rely on simulated occlusions that may not reflect real acquisition conditions, ZODIAC generalises by design to unexpected missing regions. Preliminary evaluations on volunteer data demonstrate that ZODIAC matches the supervised variant under standard conditions and can outperform it by up to 22\% in HD95 when partial observations contain unexpected missing regions. These results suggest that zero-shot generative priors are a principled alternative to paired supervision for anatomical shape completion in clinical ultrasound.

\bibliographystyle{splncs04}
\bibliography{refs}

\end{document}